\documentclass[letterpaper, 10 pt, conference]{ieeeconf}  

\IEEEoverridecommandlockouts                              

\usepackage{graphicx}
\usepackage{amsmath}
\usepackage{amssymb}

\title{\LARGE \bf
Bayesian deep learning of affordances from RGB images
}

\author{Lorenzo Mur-Labadia and Ruben Martinez-Cantin
\thanks{Lorenzo Mur-Labadia and Ruben Martinez-Cantin are with the Instituto de Investigacion en Ingenieria de Aragon (I3A), University of Zaragoza, Spain
        {\tt\small murloren@gmail.com, rmcantin@unizar.es}}%
}

\begin{document}

\maketitle
\thispagestyle{empty}
\pagestyle{empty}

\begin{abstract}

Autonomous agents, such as robots or intelligent devices, need to understand how to interact with objects and its environment. Affordances are defined as the relationships between an agent, the objects, and the possible future actions in the environment. In this paper, we present a Bayesian deep learning method to predict the affordances available in the environment directly from RGB images. Based on previous work on socially accepted affordances, our model is based on a multiscale CNN that combines local and global information from the object and the full image. However, previous works assume a deterministic model, but uncertainty quantification is fundamental for robust detection, affordance-based reason, continual learning, etc. Our Bayesian model is able to capture both the aleatoric uncertainty from the scene and the epistemic uncertainty associated with the model and previous learning process. For comparison, we estimate the uncertainty using two state-of-the-art techniques: Monte Carlo dropout and deep ensembles. We also compare different types of CNN encoders for feature extraction. We have performed several experiments on an affordance database on socially acceptable behaviours and we have shown improved performance compared with previous works. Furthermore, the uncertainty estimation is consistent with the the type of objects and scenarios. Our results show a marginal better performance of deep ensembles, compared to MC-dropout on the Brier score and the Expected Calibration Error.

\end{abstract}

\section{Introduction}

Autonomous robots need to interact with objects and its environment. A simple task for a human implies a deep understanding on how to interact with objects and how they interact between them. For example, while eating, a kid knows how to grasp a fork to use it properly, how to take the piece of food from the dish and how to take it to his mouth. Defined by the psychologist J.J Gibson \cite{gibson}, affordances are the different action possibilities available in the environment depending on the motor and sensing capabilities of the individual. They relate the objects, the actions and the possible effects of that actions carried on the objects \cite{affmontesano}. Thus, agents need to understand all the possible actions associated to an object in order to interact with it:  a cup is \textit{graspable}, a road is \textit{traversable} and a chair is \textit{sitable}, but can be also \textit{graspable}. In addition, there might be some restrictions to the actions based on the context or due to human interactions, such as social norms. For example, although it is physically plausible, we would not want for a robot to place food on the floor, or it might not use a chair that it has already been occupied. Furthermore, grasping a cup with hot tea or cross a street with cars moving can be dangerous for others or ourselves, so we should not take these actions. The concept of affordances has been widely exploited in cognitive robotics as an efficient way to represent high-level abstractions of the real world and encode information about the objects behaviour \cite{montesano2}. It can be further exploit in searching the best object to execute a task, tool usage \cite{montesano6}, predict the consequences of an action and imitation learning \cite{affmontesano}. Based on this, affordance prediction emerges as a powerful tool, with exciting and promising applications such as interaction with other agents \cite{sociable}, visual assistant devices \cite{pros_vision}, virtual reality simulations \cite{augreal} or humanoid robots.

\begin{figure}
\centering
\includegraphics[width=0.99\columnwidth]{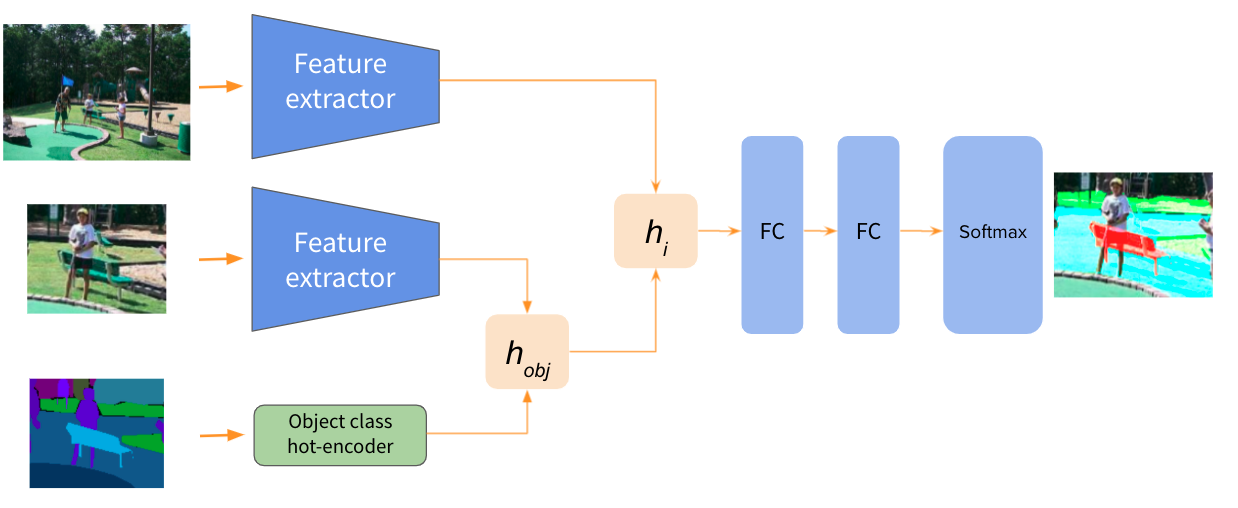}
\caption{Architecture of our model: the CNN encoder extracts the semantic information from the object and the global scene, which are combined with the object-class. Then, two fully connected layers and a softmax layer predict the affordance class.}
\label{fig:kb}
\end{figure}

Deep learning has emerged as a powerful instrument in computer vision and robotic perception to achieve scene understanding and extract relevant information from the environment. For example, using the architecture of Fig. \ref{fig:kb} we can combine different level of features with semantic segmentation to obtain an affordance based on the object and its context. Furthermore, there is a growing interest on dealing with uncertainty in deep learning \cite{bayesi, baythesis, Gal, bayesbi, bayesbib2}, specially for computer vision and image processing tasks. By measuring the level of uncertainty in the visual information of an scene, it is possible to know the level of reliability of the predictions and reduce the impact of these uncertainties during the optimization and decision making processes. However, Bayesian deep learning methods are computationally expensive, even though the posterior distribution is approximated with a Gaussian distribution using Laplace approximation or variational inference \cite{introbayes, introbayes2, introbayes3}, or by having a sampled distribution like in the case of Monte Carlo dropout \cite{Gal2016} and deep ensembles \cite{Lakshminarayanan2017}. Because we are interested in discrete predictions and the methods are more efficient, in this work we will focus on the sampling methods. Furthermore, sampling methods are also able to approximate the multi-modality of the posterior distribution. Using the posterior uncertainty we can discard low-confidence results, reason about similarities between classes, model noisy observations, analyse sources of uncertainty and also to serve as the basis for future active learning algorithms that improve our dataset \cite{active}.


The main contribution of this paper is the uncertainty estimation in the prediction of affordances comparing MC-Dropout and Ensembles. Results show that aleatoric uncertainty occurs due to the presence of noise inherent to the observations in far objects, while epistemic uncertainty appears in samples out of the distribution (e.g those objects that appear more rarely in the dataset).  Ensembles perform better than MC-Dropout in Brier Score and ECE metrics due to its capacity of capturing the multi-modality present in the posterior distribution. In addition, our model was improved by comparing the effect of the feature extractors in the encoder and the effect of dropout as regularizer. Finally, an analysis on the dataset shows that the biased and sparse distribution of the categories is reflected later on the metrics of the model. We hope that this paper serves as base and inspiration for future researches in the field of visual affordances, a fundamental step to develop future active agents.

\section{Related work}

In this section, we present some related work on affordance detection and Bayesian deep learning.

\subsubsection{Affordance reasoning} 
Multiple approaches have tried to overcome the detection of affordances using different methods. In the seminal paper from Montesano at al, the authors propose to learn affordances through unsupervised interaction of a robot with the environment using Bayesian networks dealing with uncertainty or irrelevant information \cite{affmontesano}. This Bayesian approach is extended in \cite{ruben}, where they use Gaussian Mixture Models (GMMs) in order to model the sensory data and considering explicitly the probability distribution in each affordance concept. Ardon et al. \cite{affrobotics} compare the different level of prior knowledge that it is assumed in the affordances detection. However, most succesful applications of  affordances in robotics have focused mainly on how to $grasp$ an object \cite{affgrasp, affgrasp2, affgrasp3}. In \cite{hand}, they extend this idea using an anthropomorphic robotic hand to make multi-step predictions in table cleaning and object moving applications. As affordances encode relationship between an action, object and effect, reinforcement learning approaches such as \cite{affrl} also tried to overcome the problem but in an artificial 2D environment. Deep learning methods have allowed affordance detection directly through computer vision with a deeper scene understanding \cite{affcv2,affgrasp4,aff23}. The use of latent space in affordance encoding was exploit by \cite{aff_feat} to demonstrate that they can be used to train policies successfully, as they are invariant to textures or distractor objects. Finally, \cite{actprop} collects a new dataset referred to as ADE-affordance and use Graph Neural Networks (GNNs) to propagate the contextual information. In their architecture, the hidden state of each node is updated by combining the influence of its neighbours and its own memory.

\subsubsection{Bayesian deep learning} 
As discussed in before, in this work we are going to focus on sampling based methods of Bayesian deep learning. Deep ensembles \cite{Lakshminarayanan2017}, similarly to other ensemble based methods use a randomization scheme to perform Bayesian model averaging. Therefore, deep ensembles are not \emph{truly Bayesian} in the sense that the posterior distribution is never computed. However, in practice, previous works have shown that the sampling distribution can be a good proxy for the posterior distribution. On the other hand, Gal and Ghahramani \cite{Gal2016} found that the dropout method, which randomly switches off some neural connections and was previously used as a regularization method during training, can also be interpreted as a Monte Carlo algorithm when it is performed during prediction. Therefore, it can also be used to obtain a sampled distribution of the posterior. However, each prediction requires multiple forward passes through multiple sampled models. Ensembles and MC-dropout are extensively compare in \cite{bayesi} for real-world applications in computer vision such as street scene semantic segmentation or depth completion. 

The posterior distribution obtained in both method can be further separated in aleatoric and epistemic uncertainty. Aleatoric uncertainty refers to the variations caused by the realization of different experiments with stochastic components. In our models, it encodes the variability in the different inputs from the test data and hence cannot be reduced by increasing the amount of training data. Epistemic uncertainty represents the lack of knowledge of a trained model. This type of uncertainty is deeply related to the training data and the model ability to generalize. Kendall and Gal \cite{Gal} analyzed both uncertainties in common computer vision tasks (semantic segmentation and depth regression) to show that the aleatoric and epistemic uncertainty model different phenomenons in an scene. Aleatoric uncertainty appears in the contours of objects and far away regions, as it captures the noise associated to the observation. Higher values of epistemic uncertainty were in those pixels where the segmentation model failed to predict. Kwon and Won \cite{method} improved the method presented by \cite{Gal} to decompose the sources of uncertainty using a variational inference method.  In \cite{mc2}, they compare epistemic and aleatoric uncertainty using convolutional neural networks (CNNs) applied to medical image segmentation problems at pixel and structure levels. In this case, they use MC-Dropout to estimate the distribution of the output segmentation. 

\section{BAYESIAN DEEP LEARNING OF AFFORDANCES}

 Our goal is to predict a target action-object relationship $y \in \mathbb{Y}$ given an input image $x \in \mathbb{X}$ and a type of affordance $a \in \mathbb{A}$, our neural network is a parametric function $f_w : \mathbb{X}, \mathbb{A} \rightarrow \mathbb{U}$ with parameters $w \in \mathbb{R}^P$. For the affordance problem, we can assume that the relationship type is discrete, which can be modeled as a classification problem \cite{calibration_neural} \cite{bayesi}. Our classification model can be seen in Fig. \ref{fig:kb}, which is based on the architecture from \cite{actprop}, which is also based on a semantic segmentation network. The model predicts the effect of taking an action with an object, which combines information from the object segmentation (object class and local image features) with contextual information from global image features. It uses a CNN architecture as encoder in order to extract the semantic information in a low dimensional space from the object and the global scene. In previous works, a standard pre-trained model is used as encoder and feature extractor, such as Resnet-50 \cite{actprop}, but that model might be too large for an embedded application like a mobile robot or a wearable device for assistance. In this work, we compare three models: Resnet-50 (2048 components in the latent vector space) \cite{res50}, Resnet-18 (512 components) and Mobilenet v3-small (576 components) \cite{mobilenet}. The object vector $h_{obj}$ was composed by the object class vector (one-hot encoded) $\hat{c}$ and feature vector $\phi(o)$:
 \begin{equation}
    h_{obj} = g(W_c \hat{c}) \odot g(W_f \phi(o))
    \label{eq:objvec}
\end{equation}
where $W_c$ and $W_f$ are the layer weights, representing fully connected layers that map respectively the object class and the image feature vector to the 128-components space of the hidden representation, while $g(\cdot)$ is the non-linear ReLU and $\odot$ a element-wise multiplication. Note that the object features are computed from the extended object's bounding box by a factor of 1.2 to include more contextual information. The image information vector $h_i$ can be obtained by a concatenation of the object vector $h_{obj}$ with the global features of the scene $\phi(I)$ passed by a second fully connected layer $W_h$:
\begin{equation}
    h_i = g (W_h [h_{obj} \phi(I)])
    \label{eq:imgvec}
\end{equation}
Then, the information vector activations are sent to another two fully connected (FC) layers and a softmax layer to get the affordance prediction for each action as vector $p$ with a probability for each class. Then, $\hat{y} = \arg \max p$. Training is performed using the cross-entropy loss.

\subsection{Bayesian deep learning}

Many recent work on Bayesian deep learning for computer vision use Monte Carlo dropout or deep ensembles \cite{bayesi,introbayes,introbayes2,introbayes3}.

Monte Carlo dropout (MC-droput) approximates the posterior as the sample distribution of $M$ forward passes during the test time with random dropout layers active \cite{Gal}, which is equivalent as sampling from the approximate posterior $p(w | x, y)$, which cannot be computed analytically. It can be interpreted as a special case of variational inference, where the Kullback-Leibler (KL) divergence of the approximate posterior with respect to the true posterior is minimized by approximating the variational parameters of a Bernoulli distribution \cite{Gal2016}. In practice, the Bernoulli distribution defines the dropout probability for each neural connection.

Deep ensembles method is a Bayesian model average technique that although increases linearly the training time, it works better than MC-dropout when the posterior distribution does not follow a Bernoulli distribution. In \cite{Lakshminarayanan2017}, they train $M$ different models with a random initialization of their neural network parameters, minimizing the negative log-likelihood loss $\mathcal{L}=-\log p(y | x, w)$ and shuffling the dataset, trying to combine models to discover a more powerful model. In practice, given that the number of samples of both MC-dropout and deep ensembles are quite small in comparison with the huge dimensionality of the parameter space, deep ensembles have the advantage that all the samples are \emph{optimized}. From a Bayesian point of view, the posterior approximation is incorrect, but in practice, their performance is better because we avoid spurious cases where all the sampled models have low likelihood (and therefore, low performance). Furthermore, since we are minimizing MLE starting from randomly initial point, we are finding different local optima, capturing the multi-modality of the posterior distribution even with few samples \cite{bayesi}. 

For both methods, their final prediction is the average of the predicted probabilities 
\begin{equation}
p(y | x) = \frac{1}{M} \sum_{m=1}^M p (y | \textbf{x}, w_m).
\end{equation}

\begin{table}
\caption{Categories distribution in the training dataset}
\label{tab:categdist}
\centering
\begin{tabular}{c|ccc}
                & Sit  & Run  & Grasp \\ \hline
Firmly positive & 48.3 & 12.6 & 26.5  \\ \hline
Exceptions      & 14.0 & 16.9 & 5.0   \\ \hline
Firmly negative & 37.6 & 70.5 & 68.5
\end{tabular}
\end{table}

\begin{table*}
\caption{Relationship prediction comparing different feature extractors}
\label{tab:determres}
\centering
\begin{tabular}{c|ccc|ccc|ccc}
\hline
\centering
& \multicolumn{3}{c}{Sit}   & \multicolumn{3}{c}{Run}  & \multicolumn{3}{c}{Grasp} \\
& m-Acc E & m-Acc & m-Acc B & m-Acc E & m-Acc & m-Acc B & m-Acc E  & m-Acc & m-Acc B \\ \hline
Mobilenet                                                        & 0.816   & 0.856 & 0.854   & 0.833   & 0.921 & 0.928   & 0.858    & 0.855 & 0.878   \\
\begin{tabular}[c]{@{}c@{}}Mobilenet Dropout 0.3\end{tabular} & \textbf{0.820}   & \textbf{0.860} & 0.865   & 0.834   & \textbf{0.924} & 0.929   & \textbf{0.860}    & \textbf{0.861} & \textbf{0.895}   \\ \hline
Resnet-50                                                        & 0.768   & 0.797 & 0.823   & 0.815   & 0.834 & 0.835   & 0.833    & 0.831  & 0.882    \\
\begin{tabular}[c]{@{}c@{}}Resnet-50 Dropout 0.3\end{tabular} & 0.787   & 0.802 & 0.813   & 0.796   & 0.870 & 0.930   & 0.839    & 0.830 & 0.893       \\ \hline
Resnet-18                                                        & 0.815   & 0.825 & 0.861   & \textbf{0.836}   & 0.920 & 0.923   & 0.855    & 0.860 & 0.878  \\
\begin{tabular}[c]{@{}c@{}}Resnet-18 Dropout 0.3\end{tabular}  & 0.819   & 0.857 & \textbf{0.868}   & 0.835   & 0.922 & \textbf{0.931}  & 0.859    & 0.859 & 0.885        \\ \hline

\end{tabular}
\end{table*}

The variance of the predictive distribution $p(y^* | x^*, \mathcal{D})$, with $\mathcal{D} = \{x, y\}_{j=1}^{N_d}$ being the training database, can be decomposed as the sum of the epistemic and aleatoric uncertainty. Epistemic uncertainty  $\sigma_e$, related to the model knowledge, is reduced as the sample size increases. It can be expressed as a probability distribution over the model parameters. Previous results show that in computer vision applications, epistemic uncertainty appears in where the input sample (image, pixel...) is out-of-distribution from the training data \cite{Gal}. Aleatoric uncertainty $\sigma_a$ is associated to the noise inherent in the observations such as motion noise, objects far from the camera or object boundaries and it cannot be reduced by collecting more data as it is inherent to the data distribution. In our case, both can be calculated from the the output probability vector $p_m$ for each sample model $m$ \cite{method}. The aleatoric uncertainty can be estimated as:
\begin{equation}
    \sigma_a = \frac{1}{M} \sum_{m=1}^{M} diag({p_m}) - {p_m} {p_m}^T
    \label{eq:aleatoric}
\end{equation}
with the epistemic uncertainty being
\begin{equation}
    \sigma_e = \frac{1}{M} \sum_{m=1}^{M} ( {p_m}-\widehat{p}_m)({p_m}-\widehat{p}_m)^T
    \label{eq:epistemic}
\end{equation}
and $\widehat{p}_m = \frac{1}{M} \sum_{m=1}^{M} p_m$.

\section{Experiments}

In this section, we present the training process and the experimental setup where we use a standard image dataset for affordance prediction.

\subsection{ADE-Affordance dataset}


The ADE-Affordances dataset \cite{actprop}, composed by 44,302 objects divided in 10,000 scenes, was build on top of some of the ADE20K \cite{ADE20} scenes for three different actions: \textit{sit, run} and \textit{grasp}. It is not only limited to provide information referent to a positive or negative \textit{object-action} relationship, but also it explains in 7 categories, including exceptions with social meaning (categories 1-5):
\begin{enumerate}
    \item[0-] Positive, we can take this action
    \item[1-] Object non-functional for this action
    \item[2-] Physical obstacles prevent you to take the action
    \item[3-] Socially awkward, it is not proper to take the action
    \item[4-] Socially forbidden
    \item[5-] Dangerous to ourselves or others
    \item[6-] Firmly negative, never take this action
\end{enumerate}

Previous results on this dataset showed a subpar performance using a classifier architecture, requiring the use of more convoluted models such as graph neural networks \cite{actprop}.
Our analysis shows that the number of objects in each class and the ratio of classes (Table \ref{tab:categdist}) are biased distributed. For example, most of the affordances for the \textit{run} action are firmly negative (70.7 $\%$) compared with the ratio of exceptions (16.7 $\%$) or firmly positive (12.6 $\%$). Object apparition also reflects the sparsity of the dataset, since most of the samples are concentrated in a few objects (\textit{chair, floor, pot, book}): e.g while \textit{chair} appears a 20.7 $\%$, \textit{bench} is only in a 0.96 $\%$ of the classes or \textit{towel} a 1.4 $\%$.

\begin{table*}
\caption{Relationship prediction metrics comparing Ensembles with MC-Dropout}
\label{tab:bayesres}
\centering
\begin{tabular}{c|ccc|ccc|ccc}
                              & \multicolumn{3}{c|}{Sit}   & \multicolumn{3}{c}{Run}    & \multicolumn{3}{c}{Grasp} \\
                              & m-Acc E & m-Acc  & m-Acc B & m-Acc E & m-Acc  & m-Acc B & m-Acc E & m-Acc  & m-Acc B \\ \hline
Ensembles N = 5               & 0.819  & 0.854 & 0.865  & 0.834  & 0.921 &  0.924  & 0.859  & 0.860 & 0.889 \\
Ensembles N = 10              & 0.820  & 0.854 & 0.866 & 0.834  & 0.922 & 0.925 & 0.859  & 0.860 & 0.889 \\
Ensembles N = 25              & \textbf{0.821}  & 0.855 & \textbf{0.870}  & \textbf{0.835}  & 0.922 & 0.926  & 0.859 & \textbf{0.861} & 0.890  \\
Ensembles N = 50              & \textbf{0.821}  & 0.855 & \textbf{0.870} & \textbf{0.835}  & 0.922 & 0.927  &\textbf{0.860}  & \textbf{0.861} & \textbf{0.891} \\
\hline
MC-D = 0.1 (M = 50)                 & 0.818  & \textbf{0.861} & 0.864  & 0.834  & 0.922 & \textbf{0.929}  & \textbf{0.860}  & 0.860 & 0.881 \\
MC-D = 0.3 (M = 50)                   & \textbf{0.821}  & 0.855 & 0.869  & \textbf{0.835} & \textbf{0.923} & 0.928  &\textbf{0.860} & \textbf{0.861} & 0.885 \\
MC-D = 0.5 (M = 50)                   & 0.778  & 0.845 & 0.868  & 0.780  & 0.918 & 0.926  & 0.798  & 0.855 & \textbf{0.891} \\
\end{tabular}
\end{table*}


\subsection{Model training}

We train our model using Adam \cite{adam} with batch-size of 128 samples and an initial learning rate of $10^{-4}$ that decay after 5 epochs by a factor of 0.85. Features of each object were extracted from the last hidden layer and compared from three state-of-the-art CNN feature extractors: Resnet 50 \cite{res50}, Resnet-18 and Mobilenet-v3 \cite{mobilenet}. At the Bayesian models, we train and inferred them using different dropouts rates: 0.5, 0.3 and 0.1 with $M=50$ samples. Due to their much higher training time, deep ensemble models were trained for different number of models $M = 2, 5, 10, 25, 50$.

\subsection{Metrics}

For the deterministic experiments, we used three different metrics, depending on the treatment of exceptions, that is, those effects that are not firmly positive or negative. \textit{Mean accuracy with exceptions} (mAcc-E) treats each exception as a different class (7 categories), \textit{Mean accuracy} (mAcc) considers all the exceptions in one class (3 categories), and \textit{Mean binary accuracy} (mAcc-B) considers a binary outcome, where all the exceptions are also considered negative. For the Bayesian experiments, we compute the covariance matrix of the aleatoric and epistemic uncertainty using \eqref{eq:aleatoric} and \eqref{eq:epistemic} respectively. We measure the evolution of its components and compare the value between the epistemic and aleatoric terms of the predicted class. Furthermore, we also use the \emph{Expected Calibration Error} and the \emph{Brier score}.
\paragraph{Expected Calibration Error (ECE)} We say that a model is miscalibrated when there is a difference between its confidence and accuracy. Thus, a simple metric of calibration is the difference between both. We can easily approximate this difference by partitioning the prediction into $L$ bins and computing the weighted average of the difference between accuracy and confidence for each bin. This procedure is shown to converge to the true expected difference \cite{calibration_neural}.
\begin{equation}
ECE = \sum_{l = 1}^L \frac{B_l}{M} \left| acc(B_l) - conf(B_l) \right|
\label{eq:ece}
\end{equation}
where the accuracy of a interval $B_l$ is the ratio of the predicted classes $\hat{y_i}$ which are true labels $y_i$ for the sample $i$, and it can be estimated as $acc(B_l) = \frac{1}{\lvert B_l \lvert} \sum_{i \in B_l}\textbf{1} (\hat{y_i} = y_i)$. The average confidence in each bin $B_l$ is defined as the average of the probabilities of correctness $conf(B_l) = \frac{1}{B_l} \sum_{i \in B_l} p_i$.

\begin{figure*}
\centering
\includegraphics[width=1.5\columnwidth]{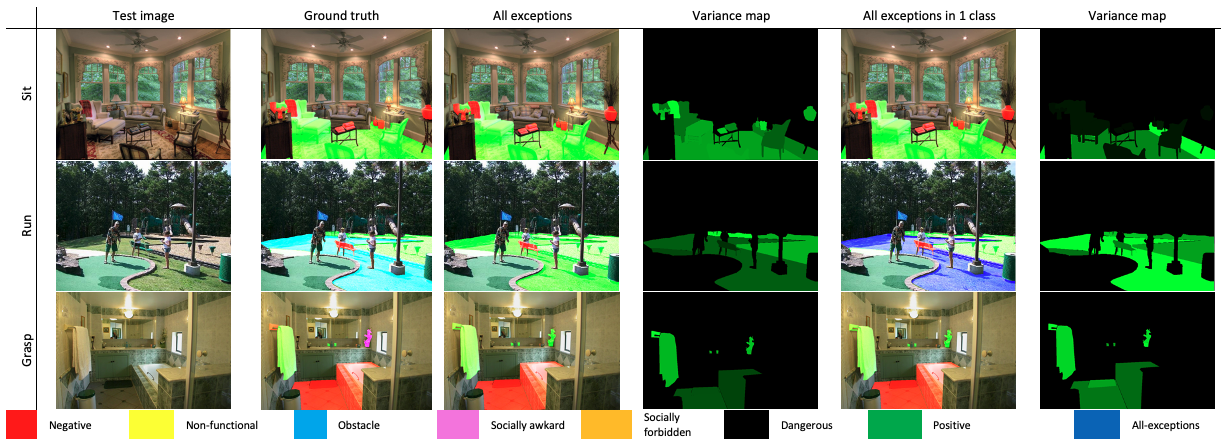}
\caption{Relationship and variance map prediction using Mobilenet features with Dropout. The brighter the green, the higher the variance}
\label{fig:segm}
\end{figure*}

\begin{figure}
\centering
\includegraphics[width=0.65\linewidth]{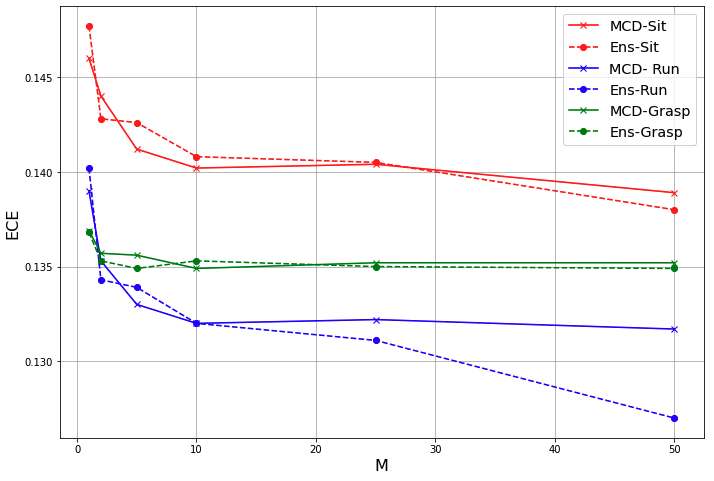}
\includegraphics[width=0.65\linewidth]{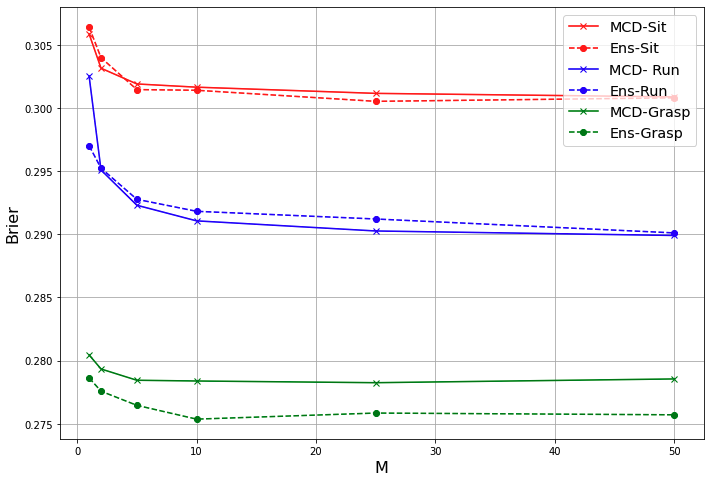}
\caption{Evolution of the ECE (top) and Brier score (down) metrics for different number of models \textit{M}}
\label{fig:metric_evol}
\end{figure}

\begin{figure*}
\centering
\includegraphics[width=0.12\linewidth]{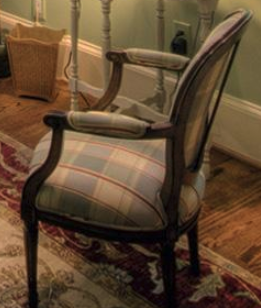}
\includegraphics[width=0.2\linewidth]{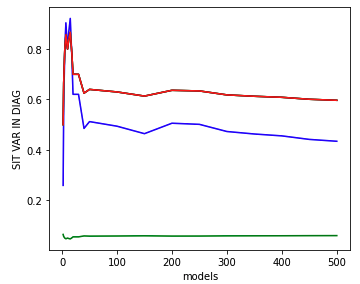}
\includegraphics[width=0.2\linewidth]{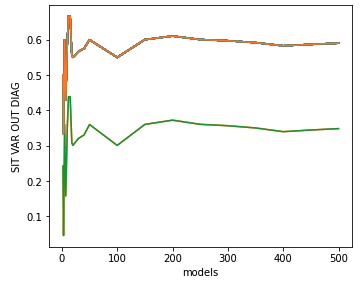}
\includegraphics[width=0.15\linewidth]{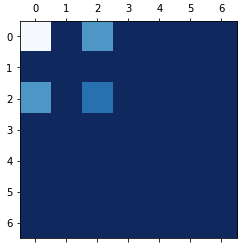}
\includegraphics[width=0.2\linewidth]{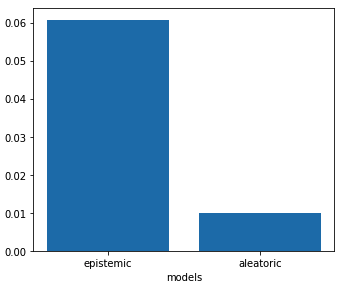}

\includegraphics[width=0.14\linewidth]{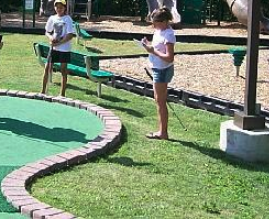}
\includegraphics[width=0.2\linewidth]{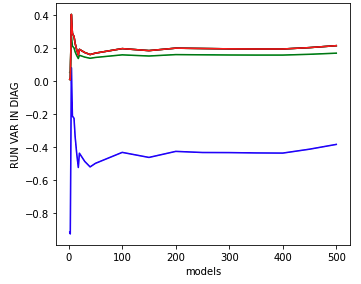}
\includegraphics[width=0.2\linewidth]{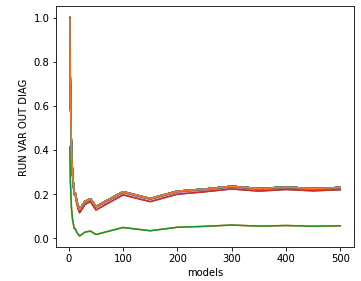}
\includegraphics[width=0.15\linewidth]{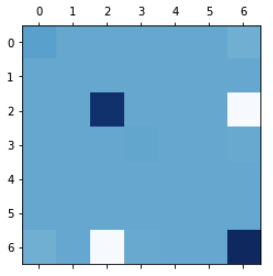}
\includegraphics[width=0.2\linewidth]{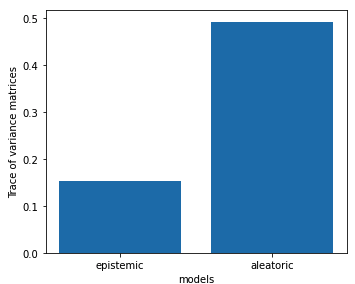}

\includegraphics[width=0.13\linewidth]{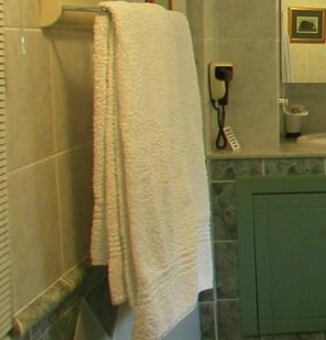}
\includegraphics[width=0.2\linewidth]{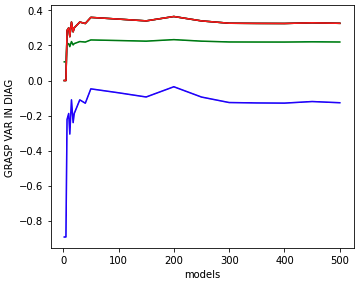}
\includegraphics[width=0.2\linewidth]{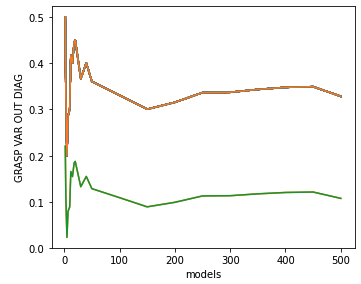}
\includegraphics[width=0.15\linewidth]{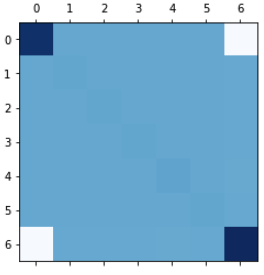}
\includegraphics[width=0.2\linewidth]{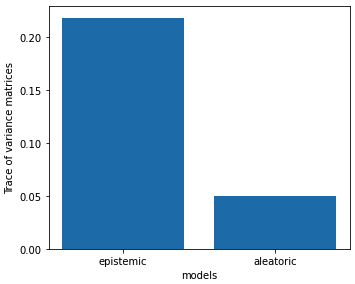}

\includegraphics[width=0.95\linewidth]{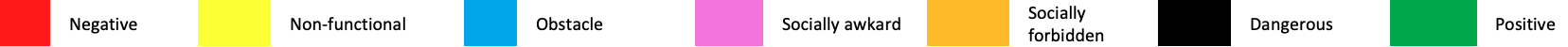}

\caption{From left to right: A) Image of the object. B) Evolution of the components in the trace of the covariance matrix. C) Evolution of the components out the trace of the covariance matrix. D) Covariance matrix for the model $M$=500. E) Value of the epistemic and aleatoric terms of the predicted class}
\label{fig:var_evol}
\end{figure*}

\paragraph{Brier Score (BS)} This metric was stated in 1950 by Glenn W.Brier \cite{brier} in order to measure the accuracy of weather forecast predictions. The multiclass formulation can be expressed as
\begin{equation}
BS = \frac{1}{M} \sum_{m=1}^{M} \sum_{i=1}^{R} (p_{mi} - \widehat{c}_{mi})^2
\label{eq:brier}
\end{equation}
where $R$ the number of classes and $\widehat{c}$ is the one-hot encoding of the class (1 for the true class, and 0 for the rest). A perfect accurate model obtains a Brier score of 0, while a 1 means that the model is completely inaccurate.

\section{Results}

We have conducted an experiment using a deterministic model with different encoder. Then, the optimal encoder was used in a Bayesian experiment using MC-dropout and deep ensembles.

\subsection{Deterministic model}
We compared Resnet-50 that outputs 2048 feature components with two lighter feature extractors such as Resnet-18 and Mobilenet v3-small, which output features of 512 and 576 components respectively. As Table \ref{tab:determres} shows, a smaller model size implies a faster training of the model but also achieves better results. Mobilenet converged in the first 20 epochs while Resnet-50 took up to 200 epochs to show any improvement. 
Furthermore, adding intermediate dropout layers before each fully connected layer help to prevent over-fitting, which used to occur at the mAcc and mAcc-B models. A dropout rate of 0.3 show the better performance predicting the affordances. Results for the relationship prediction can be seen in Fig. \ref{fig:segm}, where the model correctly predicts affordances from the object classification and the contextual information from the scene. It distinguishes properly between positive and firmly negative classes, although in some cases it fails classifying the type of exception. Note that in this case, dropout is only applied during training as a regularizer. The model during evaluation is purely deterministic.

\subsection{Bayesian experiment}
Based on the deterministic experiment, we use Mobilenet-v3 as the feature extractor for the Bayesian experiments. We trained the same architecture for a different number of deep ensembles (up to M=50) and different MC-dropout rate. The performance of MC-Dropout on the accuracy metric is similar to deep ensembles, as shows Table \ref{tab:bayesres}; and Bayesian models slightly improve the accuracy results compared to the deterministic model in Table \ref{tab:determres}. We settled 0.3 as the better dropout rate, since excessive zeroes of the neurons during testing decrease the capacity of the model and this magnitude was enough to control the overfitting on the mAcc-B.

The MC-dropout and deep ensembles models are also compared in terms of Brier Score and ECE, as shows Fig. \ref{fig:metric_evol} for the different actions. In all of the cases, these two metrics improve up to $M = 10$, followed by a distinct plateau from $M > 25$. This is also coherent with the accuracy results in Table \ref{tab:bayesres}, where the  deep ensembles performance have a diminishing return from $M = 25$. Ensembles models presents better ECE and Brier score than MC-dropout. There is also a significant difference for the three different actions: \textit{sit} is worse calibrated than \textit{run} and \textit{grass} (higher ECE value), as well as its accuracy performance is also the worst.

We also performed an extensive experiment with MC-dropout with $M=500$ to illustrate the convergence of the aleatoric and epistemic uncertainty from equations \eqref{eq:aleatoric} and \eqref{eq:epistemic} respectively. Note that we used MC-dropout because 500 deep ensembles models are intractable. The results are in  Fig. \ref{fig:var_evol}. The evolution of the components of the covariance matrix show that they converge in probability to the analytical expression as the number of models increase \cite{method}. Components in the trace of the covariance matrix reflect the variance of that category, while components out of the trace show inter-relationship between categories. It also shows correlations between the different classes and adds a new level of reasoning. For example, in the minigolf example in Fig. \ref{fig:var_evol} the model fails to predict that in the \textit{grass} object there is a \textit{physical obstacle} exception, the components of the covariance matrix show that the model truly doubts between these two categories. Aleatoric uncertainty is significant in far and blur objects where camera noise is translated to pixels level, as shows Fig.  \ref{fig:var_evol} for the \textit{grass} object. Significant epistemic uncertainty shows that the sample is out of the distribution, like the towel Fig. \ref{fig:var_evol}, suggesting that the biased and sparse distribution of the samples in the training dataset is reflected in the epistemic uncertainty. 

\section{Conclusions}
We propose a Bayesian deep learning model for affordance prediction directly from image data. Compared to previous approaches, this model has the flexibility and performance of recent deep learning method \cite{actprop} while enabling uncertainty based reasoning and robustness from Bayesian methods \cite{affmontesano}. First, we have evaluated multiple encoders showing that the subpar performance illustrated by previous approaches based on CNNs were probably due to the lack of unbiased data with respect to the model size. This effect is more clear when a dropout regulizer is included. Then, we have implemented two Bayesian extensions to the model, using MC-dropout and deep ensembles and we have compared both models in terms of accuracy and calibration, both in terms of aleatoric and epistemic uncertainty. Epistemic uncertainty appears in those samples out of the distribution, which corresponds to those objects that appear less in the dataset, while aleatoric is caused by noisy observations, far-blur objects and occlusions. The high accuracy performance in the prediction of affordances shows that the agent understand under which circumstances take an action, and the consequences of doing that, however it used to fail when detecting the type of exception but a deeper analysis of the covariance matrices shows that the model truly doubts between similar classes and the predominance of aleatoric over epistemic uncertainty. Our results show that the deep ensembles predictions are more reliable and well calibrated, which is consistent with the literature in other computer vision tasks. However, deep ensembles are much more expensive to train, scaling linearly in the number of models while MC-dropout training cost remains constant.

{\small
\bibliographystyle{IEEEtran}
\bibliography{mybooks}
}

\end{document}